%% file: aaai23.tex
\documentclass[letterpaper]{article} % DO NOT CHANGE THIS
\usepackage{aaai23}  % DO NOT CHANGE THIS
\usepackage{times}  % DO NOT CHANGE THIS
\usepackage{helvet}  % DO NOT CHANGE THIS
\usepackage{courier}  % DO NOT CHANGE THIS
\usepackage[hyphens]{url}  % DO NOT CHANGE THIS
\usepackage{graphicx} % DO NOT CHANGE THIS
\usepackage{natbib}  % DO NOT CHANGE THIS AND DO NOT ADD ANY OPTIONS TO IT
\usepackage{caption} % DO NOT CHANGE THIS AND DO NOT ADD ANY OPTIONS TO IT
\usepackage{algorithm}
\usepackage{algorithmic}
\usepackage{subcaption}
\usepackage{svg}
\usepackage[textsize=scriptsize]{todonotes}
\usepackage{amsmath}
\usepackage{multirow}
\usepackage{newfloat}
\usepackage{listings}
\DeclareCaptionStyle{ruled}{labelfont=normalfont,labelsep=colon,strut=off} % DO NOT CHANGE THIS
\floatstyle{ruled}
\newfloat{listing}{tb}{lst}{}
\floatname{listing}{Listing}
\title{Detecting and Grounding Important Characters in Visual Stories}
\author{
    Danyang Liu,
    Frank Keller
}
\affiliations{
    Institute for Language, Cognition and Computation\\
    School of Informatics, University of Edinburgh\\
    danyang.liu@ed.ac.uk, keller@inf.ed.ac.uk
}

\usepackage{bibentry}
\begin{document}

\maketitle

\begin{abstract}
Characters are essential to the plot of any story.
Establishing the characters before writing a story can improve the clarity of the plot and the overall flow of the narrative.
However, previous work on visual storytelling tends to focus on detecting objects in images and discovering relationships between them. In this approach, characters are not distinguished from other objects when they are fed into the generation pipeline. The result is a coherent sequence of events rather than a character-centric story.
In order to address this limitation, we introduce the VIST-Character dataset, which provides rich character-centric annotations, including visual and textual co-reference chains and importance ratings for characters.
Based on this dataset, we propose two new tasks: important character detection and character grounding in visual stories.
For both tasks, we develop simple, unsupervised models based on distributional similarity and pre-trained vision-and-language models.
Our new dataset, together with these models, can serve as the foundation for subsequent work on analysing and generating stories from a character-centric perspective.

\end{abstract}

\input{1_intro}

\input{2_dataset}
\input{3_method}
% \input{4_experiments}
% \input{4_new_experiments}
% \input{4_spanbert_experiments}
\input{4_b3_experiments}

\input{5_related}

\input{6_conclusion}

\section*{Acknowledgements}
This work was supported in part by the UKRI Centre for Doctoral Training in Natural Language Processing, funded by the UKRI (grant EP/S022481/1) and the University of Edinburgh, School of Informatics and School of Philosophy, Psychology and Language Sciences.
% Use \bibliography{yourbibfile} instead or the References section will not appear in your paper
%\newpage
\bibliography{aaai23}

\newpage
\section{Appendix}
\subsection{Output of an Existing Visual Storytelling Model}
\begin{figure*}[htbp]
    \centering
    \includegraphics[width=\textwidth]{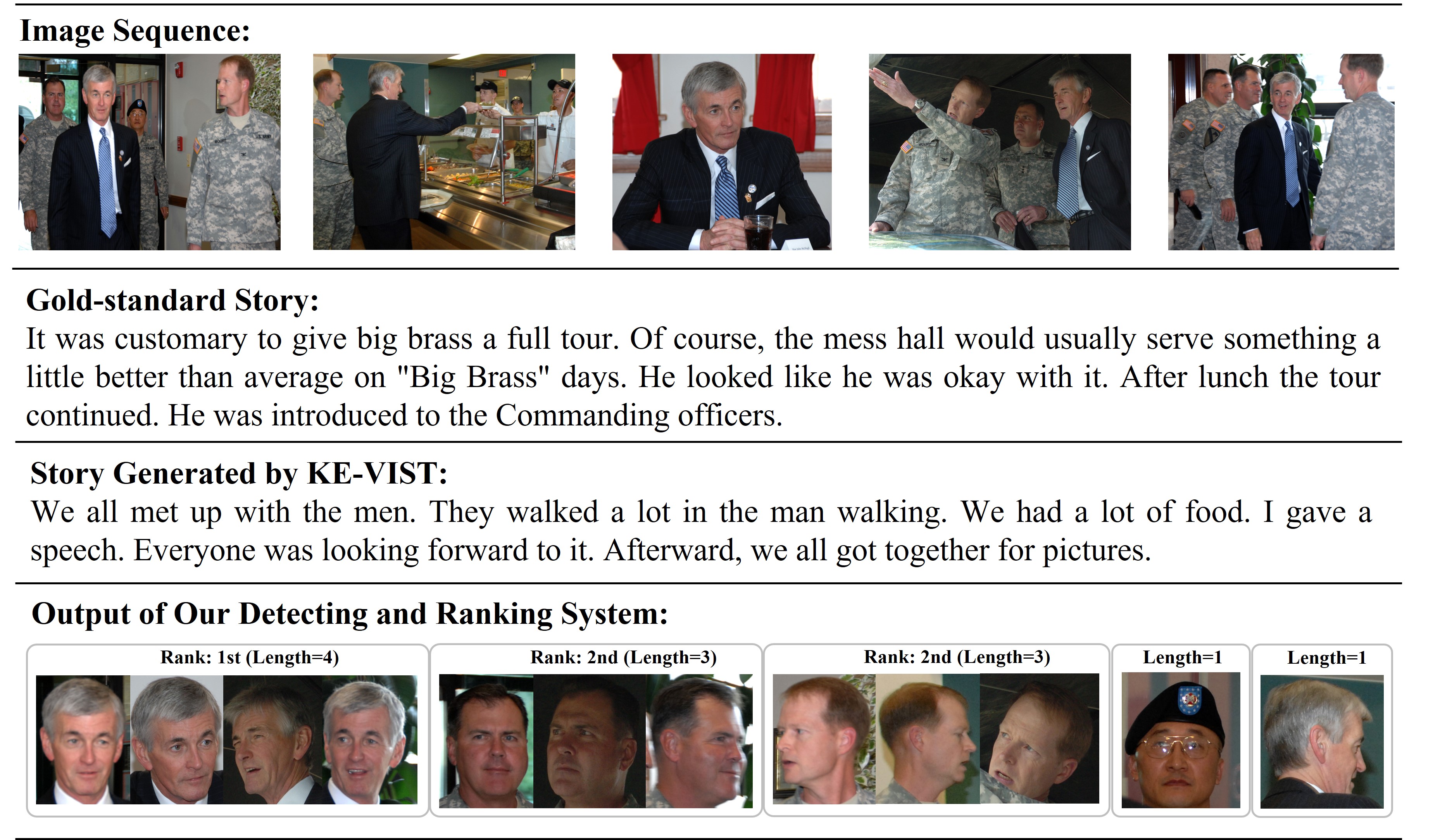}
    \caption{A representative example generated by the KE-VIST model \cite{hsu2020knowledge}, along with the gold-standard story and the corresponding output of our detecting and ranking system.}
    \label{fig:ke_result}
\end{figure*}

Existing visual storytelling models are unable to model characters and therefore create stories with false co-reference and arbitrary characters.
Figure \ref{fig:ke_result} shows a representative example generated by the KE-VIST model \cite{hsu2020knowledge}, along with the gold-standard story and the corresponding output of our detecting and ranking system.
% \vspace{-.1cm}
\begin{enumerate}
    \item From the image sequence and gold-standard story we can see that there is an obvious protagonist.
    The KE-VIST model fails to recognize the protagonist in the third image, and recognizes it as a new character (i.e. the men vs. I), causing the story confusing and illogical.
    \item Our model can identify the important characters from the image sequence, which makes it possible to train character-centric visual storytelling models to fix the above issues.
    % \item As shown in the output of our system, the protagonist (the man in black suit) in the second image is not detected correctly. Because our face detector (i.e., MTCNN \cite{xiang2017joint}) does not perform well on side face recognition.
\end{enumerate}

\subsection{Group Word List}
Group nouns (e.g. team, family), also known as collective nouns, are the words that refer to a group of people but whose PoS is not \textit{NNS} or \textit{NNPS}.
We create a vocabulary list to detect such group characters. The full list is shown in Table~\ref{tab:group}.

\begin{table}[h]
\centering
\begin{tabular}{l}
\hline
Group Words \\ \hline
team, class, club, crowd, gang, family \\
government, committee,  police, couple \\
squad, each, anyone, everybody, else\\ \hline
\end{tabular}
\caption{The vocabulary list we created for group character detection.}
\label{tab:group}
\end{table}

\subsection{Cases and Analysis of CLIP-based Grounding}
\begin{figure*}[htbp]
    \centering
    \vspace{-1cm}
    \includegraphics[width=\textwidth]{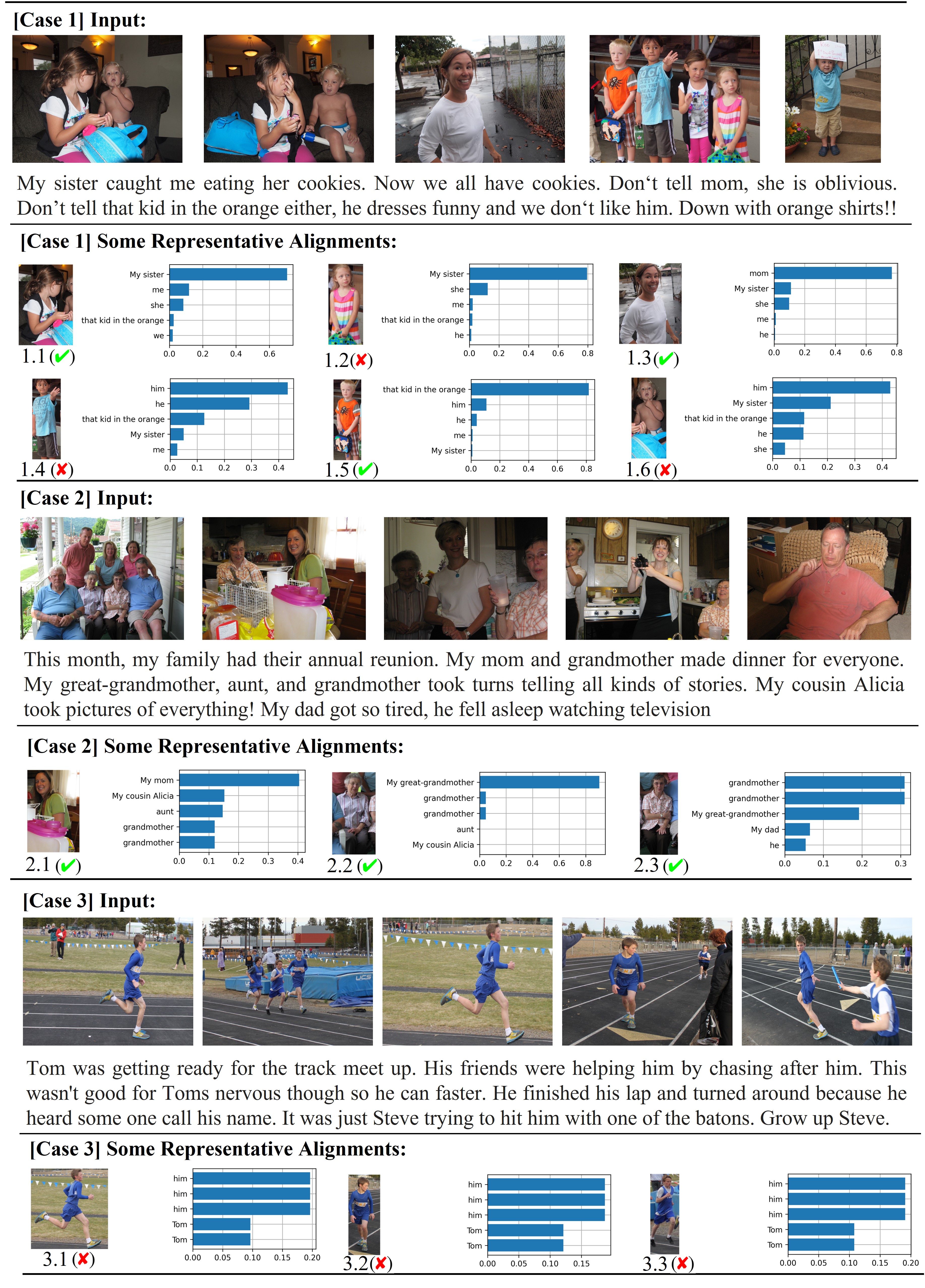}
    \caption{Some representative cases of CLIP-based character grounding with gold-standard annotations as input. Each chart shows the top-5 mentions that best match that character image.}
    \label{fig:clip}
\end{figure*}

We experiment with CLIP, a large pre-trained vision-and-language model, for the character grounding task. More specifically, we use CLIP to compute the similarity between textual and visual coreference chains by taking the average of CLIP-based similarity between each pair of textual mention and visual appearance. Figure \ref{fig:clip} shows examples of CLIP-based grounding results with gold-standard annotations as input.
Note that we only show some representative alignments. The figure shows the similarity between each pair of textual mention and visual appearance. The results will be averaged and fed into the Kuhn–Munkres algorithm \cite{munkres1957algorithms} to obtain the final chain-to-chain alignments.
From Figure \ref{fig:clip}, we can see that:

\begin{enumerate}
    \item \textbf{CLIP can distinguish between different genders, ages, and colours.} From 1.1, 1.2 and 1.3 in Case 1, we can learn that CLIP correctly matched \textit{mom} and \textit{sister} with the corresponding visual appearances.
    Case 2 further confirms CLIP's ability to identify ages by successfully grounding \textit{mom}, \textit{grandmother}, and \textit{great-grandmother}.
     If we compare 1.4 and 1.5, we can also see that CLIP can recognize colors.
    \item  \textbf{CLIP cannot handle pronouns and generic words.}
    % \textbf{Pronouns and generic words are a big problem for CLIP-based grounding.}
    CLIP-based grounding depends on the semantics of phrases. However, pronouns (e.g., \textit{he}, \textit{him} and \textit{me} in Case 1) and generic words (e.g., \textit{men}, \textit{women}) can only express basic characteristics, i.e., gender, age, and singular-plural, etc. Therefore CLIP suffers from grounding pronouns and generic words when most characters are similar in terms of these basic characteristics.
    The fact is that pronouns and generic words appear frequently in the textual co-reference chains in the VIST stories, which makes it difficult for CLIP to predict the alignment relation. Case 3 illustrates more clearly that the problem of grounding pronouns (\textit{him}, \textit{he}) and generic words (\textit{Tom}, \textit{Steve}) is amplified when the characters are all of the same gender and similar age.
    \item \textbf{CLIP prefers specific and detailed phrases.} CLIP is pre-trained on image-caption pairs whose caption is usually specific and detailed. The model can make more accurate predictions if there is more information in the phrase, e.g.,  \textit{that kid in the orange shirt} in Case 1.
\end{enumerate}

% \section{Comparison between Clustering Algorithms}
% We experiment with different clustering algorithms, including those that require a predefined number of clusters and those not.

\subsection{Ethical Consideration}
As mentioned in the submission, we augment the test set of the VIST dataset \cite{huang2016visual} which is public for academic use and does not contain sensitive information.
We obtain research ethical approval from our institute.
We will make the dataset and the codebase of the project freely available online for academic use without copyright restrictions.

Our data construction involves human participation.
Four annotators with linguistic and NLP backgrounds were recruited and trained.
They were asked to annotate character-related information and filter out samples that might cause ethical problems.
No sensitive personal information is involved in the process.
We set an appropriate salary for them.
They were also paid during the training process.

\end{document}

%% file: 1_intro.tex
\section{Introduction}
\label{sec:intro}

% the significance of the task
% 是新任务，所以花点篇幅介绍任务的重要性和意义

% VIST是一个重要任务，输入是图片序列，输出是一个连贯有意义的故事。
Visual storytelling, which aims to generate a fluent and coherent story based on an input image sequence, has recently received increasing attention from both Computer Vision and Natural Language Processing areas \cite{hsu2021plot,xu2021imagine,chen2021commonsense,hsu2020knowledge,wang2020storytelling,huang2016visual}.
% 传统相比
Unlike the traditional image captioning task where the output is usually descriptive text, visual storytelling requires more than just knowing the objects in the image sequence and describing them literally.
A convincing narrative must be composed by connecting different images and reasoning about what happens in the story.
In general, the most important elements of a story are the plot and the characters.
A human writer usually identifies the characters in the story from the images first, and determines who the main character is based on the plot they want to describe.
Establishing the characters before writing the story will make the plot clearer and the overall narrative more convincing.
In recent years, there has been some work published focused on character-centred story generation.
For example, \citet{inoue2022learning} and \citet{liu2020character} propose to learn character representations to build a character-centric storytelling model.
\citet{brahman-etal-2021-characters-tell} present a new dataset of literary pieces paired with descriptions of characters that appear in them for character-centric narrative understanding.
There is also a long line of work which aims to link character mentions in movie or TV scripts with their visual tracks \cite{rohrbach2017generating,bojanowski2013finding,everingham2006hello,sivic2009you,tapaswi2012knock}.

% existing methods, sota
% 然而，在视觉故事讲述领域，大部分模型单纯的将人物与其他objects视为同等的线索，输入到深度网络中，生成故事。
Existing work on visual storytelling, however, tends to focus on detecting objects in images and discovering relationships between them, while neglecting character information.
Characters are treated as equally important as other objects, and fed into a generation pipeline without distinguishing them.
For example, a recent approach to visual storytelling \cite{xu2021imagine,chen2021commonsense,hsu2020knowledge, yang2019knowledgeable} is to exploit an external commonsense knowledge graph to enrich the detected objects and thus improve coherence of the generated stories.
Despite being successful in describing the sequence of events, such an approach is unable to model the characters in a story.
% 这么做的问题是这会导致生成的故事中有随机的角色。
It ends up generating stories with incorrect co-reference and arbitrary characters -- characters that don't have a consistent personality or background story.
Detailed examples of the output of an existing system can be found in the supplementary material.
% \todo{It would be great if we had an example for this.}
% 换句话说，角色之间没有主次之分，而且同一个角色没有一致的性格或者背景故事，使得故事并不可信。

Moreover, existing approaches are unable to take into account the importance of characters, which means they cannot distinguish protagonists from side characters in the generated stories. The fact that there are no visual story datasets with character annotation makes it difficult to develop models for character-centric storytelling.
% \begin{figure}[t!]
%     \centering
%     \includegraphics[width=.47\textwidth]{imgs/ke_example.jpg}
%     \caption{A story of generated by KE-VIST model \cite{hsu2020knowledge} with arbitrary characters.}
%     \label{fig:example}
% \end{figure}

% our solution, how to handle the challenges and limitations (图文两开花) (in this paper)
% might use a figure to show the story generation with/without main character detection
% 所以，我们标注了一个富有人物信息的数据集，vist-char，其中蕴含了人物在文本和图像序列中的co-ref关系，人物跨模态的对齐关系，还有人物的重要性关系。
In this paper, we introduce the VIST-Character dataset, which augments the test set of VIST (the Visual Storytelling dataset; \citealt{huang2016visual}) with rich character-related annotation.
In VIST-Character, all mentions of the characters in a story are marked in the text and identified by bounding boxes in the image sequence. Mentions of the same character (both textual and visual) are annotated as part of a single co-reference chain. Furthermore, the dataset includes a rating of all characters according to their importance in the story. This makes it possible to distinguish protagonists and side characters.
Figure~\ref{fig:example} shows an example story from the VIST-Character dataset.
%基于这个数据集之上，我们提出了一个新的任务，重要人物检测和grounding。

Based on our VIST-Character dataset, we propose two new tasks: (1) important character detection and (2) character grounding in visual stories.
The goal of the first task is to identify and rank characters according to their importance to the story. For this task, we can use the text of the story, or the sequence of images that illustrates the story, or both modalities. Important character detection can be seen as a the first step in a generation pipeline that aims to produce character-centric (rather than object-centric) stories.

The second task, character grounding, requires us to ground the textual mentions of the characters to the relevant bounding boxes in the image sequence.
This presents two challenges.
Firstly, identifying mentions of the same characters in an image sequence is harder than in videos. This is because we only have only a few discrete and discontinuous images to work with, compared to the many frames typically available for video input.
Secondly, character grounding is harder than traditional language grounding. 
% 不同与之前的grouding 任务，这个任务是chain对chain的，更加的具有挑战。
We now need to align a chain of multiple mentions of characters with a chain of multiple visual appearances. Grounding is now many-to-many, compared to traditional one-to-one ground of a single phrase to a single bounding box.

In this paper, we develop simple, unsupervised models for both important character detection and character grounding, using distributional similarity heuristics or large pre-trained vision-and-language models. We believe that our new dataset, together with these models, will serve as the foundation for subsequent work on visual storytelling from a character-centric perspective.
% 输入可以是文本，图片，或者是二者一起。对于这三种setting我们都提出了baseline模型。
% 我们相信这个数据集和任务会有助于以人物为中心的故事理解和生成的许多下游任务。
%We believe our new dataset and tasks can help not only visual storytelling but also many other downstream tasks in filed of story understanding and generation from a character-centred perspective.

%% file: 2_dataset.tex
\section{The VIST-Character Dataset}
\label{dataset}
\begin{figure*}[t!]
    \centering
    \includegraphics[width=\textwidth]{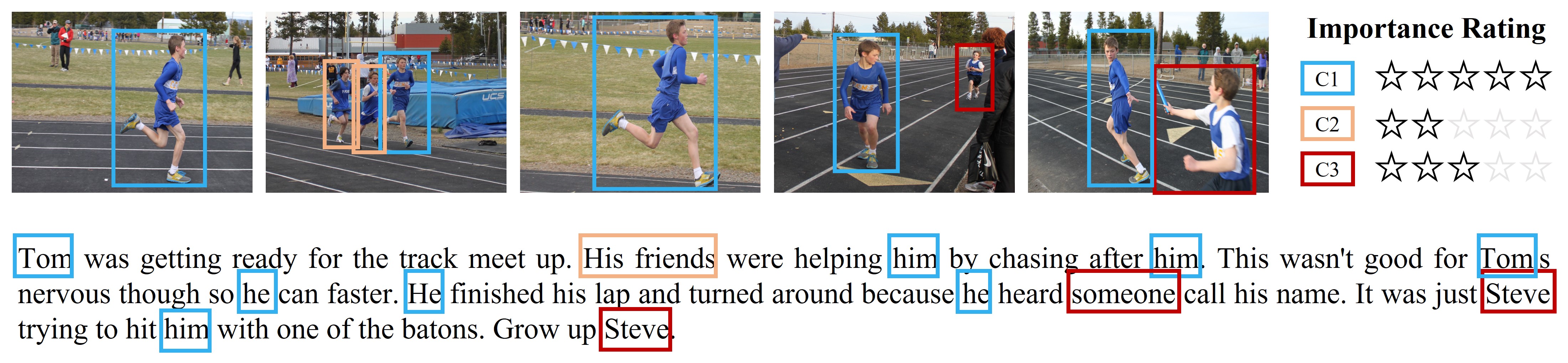}
    \caption{An example of the VIST-Character dataset. The right-hand panel shows the annotator's importance rating for each character on a scale of 1--5. The characters are colour-coded across sentences, images, and importance rating.}
    \label{fig:example}
\end{figure*}

The VIST-Character dataset contains 770 visual stories with character-related annotations, including the character co-reference chains in both stories and image sequences, and the importance rating of characters.
The visual stories are selected from the VIST test set.
We exclude the stories where there is no such character in any of the images (e.g., if the images all show landscape or scenery).
Detailed selection criteria are included in the  supplementary material.
Figure \ref{fig:example} shows an example story from VIST-Character.

\paragraph{Annotation Procedure}
The two authors conducted an annotation pilot study, based on which they devised the annotation instructions.
Overall, the annotation involves three steps: (1) Mark the words that refer to the same characters in the story sentences.
%Note that the standard version of the VIST stories consists of sentences with all character names replaced with generic “[MALE]” or “[FEMALE]” tokens.\todo{This may be too much detail. We can just say we used the original version.} Along with the transformed descriptions, the VIST dataset provides stories with the original character names. We rely on these in the annotation procedure.
(2) Identify the characters in the images. The annotators are asked to draw bounding boxes around the whole body of individual characters. For plural (e.g., \textit{students}) and group (e.g., \textit{the team}) characters, draw bounding boxes for each individual character when it refers to less than five characters, and draw a single bounding box to contain all the characters when it refers to at least five characters. (3) Rate the importance of each character by giving 1--5 stars.
We quadruple annotated fifty stories; the rest of the dataset was single annotated.
We created an annotation interface based on LabelStudio (https://labelstud.io/), an HTML-based tool that allows us to combine text, image, and importance annotation.

\paragraph{Inter-annotator Agreement}
\begin{table}[htbp]
\centering
\begin{tabular}{lrrr}
\hline
Name                         & Precision & Recall \\ \hline
Character Detection     & 76.4\%    & 71.5\% \\ %\hline
Co-reference (B-Cubed) & 82.0\%    & 79.4\% \\ %\hline
Co-reference (Exact Match)   & 63.2\%    & 58.1\% \\ %\hline
Bounding Boxes               & 67.8\%    & 58.6\% \\ %\hline
Importance Ranking           & --         & 73.0\% \\ \hline
\end{tabular}
\caption{Inter-annotator agreement for the double annotations on fifty examples of the VIST-Character dataset.}
\label{tab:double-ann}
\end{table}
Four annotators with linguistic and NLP background were trained using our annotation instructions.
Given the complexity of the task, we computed the inter-annotator agreements between the four annotators in four categories: character detection, co-reference chains, bounding boxes, and importance ranking.
Table \ref{tab:double-ann} shows the results. 
The annotation of one of the authors was used as ground truth to compute the precision and recall of of the other three annotators.
We evaluate the co-reference chains using B-Cubed and exact match.
B-Cubed is the proportion of correctly predicted mentions in a co-reference chain. 
Exact match means the predicted co-reference chain as a whole is equivalent to the gold-standard chain.
For bounding boxes, we define $IoU=\frac{|B_1\cap B_2|}{|B_1\cup B_2|}$, i.e., the intersection over the union of the bounding boxes $B_1$ and $B_2$. A bounding box is considered correct if its $IoU$ with the gold bounding box is higher than 60\%.
The results in Table~\ref{tab:double-ann} show good agreement given the complexity of the task.
 
\paragraph{Dataset Statistics}
Table \ref{tab:stat} presents various statistics of the VIST-Character dataset.
% , and Figure \ref{dist.} shows some important distributions.
We can observe that there are four characters on average in each story, and each character appears about twice.
There are 768 plural and group characters, accounting for 24.6\% of the total. This shows that plural and group characters are frequent and important in stories.
%%% removed to save space
%For the fifty instances with multiple annotations, the author checked them one by one to define the gold standard and integrate them into the final dataset.

\begin{table}[t]
\centering
\begin{tabular}{lr}
\hline
Name                               & Value \\ \hline
Number of stories                  & 770   \\ %\hline
Number of characters          & 3,119  \\ %\hline
Number of plural and group characters         & 768   \\ %\hline
Number of bounding boxes              & 4,979 \\
Average number of bounding boxes per story           & 6.47  \\ %\hline
Average number of characters per story           & 4.05  \\ %\hline
Average length of textual co-reference chain & 2.00  \\ %\hline
Average length of visual co-reference chain  & 2.02  \\ \hline
\end{tabular}
\caption{Statistics of the VIST-Character dataset.}
\label{tab:stat}
\end{table}

% \begin{figure}[htbp]
%     \centering
%     \begin{subfigure}{.5\textwidth}
%     \centering
%     \includesvg[width=\textwidth]{imgs/n_characters.svg}
%     \end{subfigure}%
%     \begin{subfigure}{.5\textwidth}
%     \centering
%     \includesvg[width=\textwidth]{imgs/length_coref_chain.svg}
%     \end{subfigure}%
%     % \begin{subfigure}{.33\textwidth}
%     % \centering
%     % \includesvg[width=1\textwidth]{imgs/n_bboxes.svg}
%     % \end{subfigure}%
%     \caption{The distribution of the number of characters (a), length of co-reference chains (b), and number of bounding boxes per character (c).}
%     \label{dist.}
% \end{figure}

%% file: 3_method.tex
\section{Task Formulation}
We propose the task of important character detection and grounding in visual stories.
The input can be a story of $C$ sentences $\{S_i\}_{i=1}^{C}$, a sequence of $C$ images $\{I_i\}_{i=1}^{C}$, or a sequence of $C$ images $\{I_i\}_{i=1}^{C}$ and associated sentences $\{S_i\}_{i=1}^{C}$ ($C=5$ in the VIST dataset). The task is then:
\begin{enumerate}
    \item For text-only input, we need to identify the character co-reference chains $\{\mathcal{T}_i\}_{i=1}^{K_t}$ in the story text, where $K_t$ denotes the the number of characters in the story text and $\mathcal{T}_i$ is the co-reference chain that contains all the mentions referring to the $i$-th textual character.
    \item For image-only input, we detect all the characters in the image sequence and cluster them into visual co-reference chains $\{\mathcal{V}_i\}_{i=1}^{K_v}$, where $K_v$ denotes the the number of characters in the image sequence and $\mathcal{V}_i$ is the visual co-reference chain that contains all the visual appearances referring to the $i$-th visual character.
    \item For multi-modal input, after the above two steps, the textual and visual co-reference chains (i.e., $\{\mathcal{T}_i\}_{i=1}^{K_t}$ and $\{\mathcal{V}_i\}_{i=1}^{K_v}$) are aligned to obtain multi-modal co-reference chains $\{\mathcal{M}_k\}_{k=1}^{K_m}$, where $K_m=\max{(K_t, K_v)}$ is the total number of characters and $\mathcal{M}_i$ is the multi-modal co-reference chain that contains all textual mentions and visual appearances of the $i$-th character.
\end{enumerate}
Regardless of the modality of the input, we also need to output an importance score for each character (i.e., for each co-reference chain).

\section{Methodology}

\begin{figure*}[htbp]
    \centering
    \includegraphics[width=\textwidth]{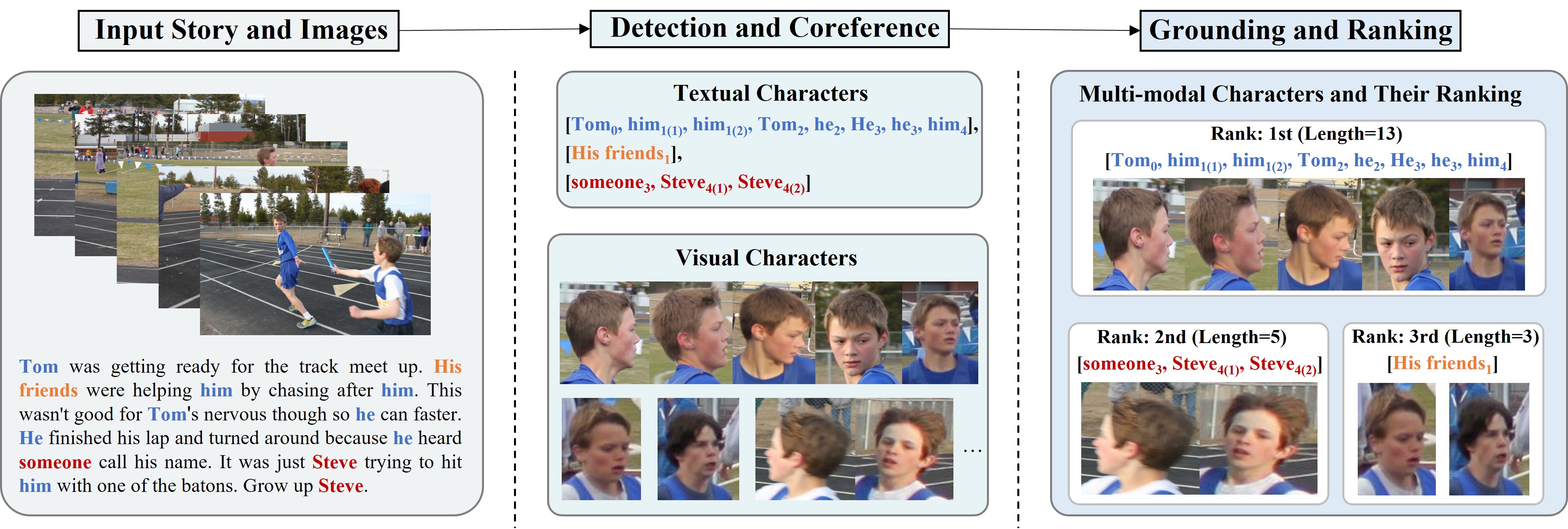}
    \caption{The architecture of the proposed model for important character detection and grounding in visual stories. Textual characters are color-coded. The subscripts of textual mentions represent their position in the story to distinguish identical words. Background characters in the images have been omitted in the figure for simplification.}
    \label{fig:arch}
\end{figure*}

In this section, we will introduce our model for detecting and grounding important characters. This model is meant as a baseline for our proposed new dataset and task.
For modelling purposes, we decompose the task into four sub-tasks: (1) character detection and co-reference resolution in the sentences, (2) character detection and co-reference in the image sequence, (3) alignment of textual and visual co-reference chains, and (4) ranking of the importance of each character.
Figure \ref{fig:arch} illustrates the proposed model architecture.
Next we will introduce the four parts respectively.

\subsection{Chraracter Detection and Co-reference in Text}
\label{sec:text-char-method}
Given a textual story $\{\mathcal{S}_i\}_{i=1}^{C}$, we find all the mentions that refer to the same character as follows:
\begin{enumerate}
    \item Use a pre-trained part-of-speech tagger to identify all noun and pronoun words in the sentences.
    \item Filter these nouns using hypernyms in WordNet \cite{miller1995wordnet}. As the VIST stories contain people, animals and vehicles as characters, we used the following hypernyms: \textit{\{person.n.01, animal.n.01, vehicle.n.01\}.}
    \item Use pre-trained co-reference resolution tools to group the mentions from step~2 into co-reference chains.
\end{enumerate}
After these steps, we have the character co-reference chains $\{\mathcal{T}_i\}_{i=1}^{K_t}$, where $\mathcal{T}_i$ denotes the co-reference chain of the $i$-th character in the story and $K_t$ denotes the total number of textual characters detected in the input sentences.

\paragraph{Plural and Group Characters}
We separately detect the mentions for plural and group character.
For plural characters, we consider words with the PoS tag \textit{NNS} or \textit{NNPS}.
For group nouns, we create a vocabulary list that contains words whose PoS is not \textit{NNS} or \textit{NNPS}, but refer to a group of people (e.g., team, family).
The full list is in the supplementary material.
We add special labels to the chains referring to plural and group characters for the subsequent grounding step.

\subsection{Character Detection and Co-reference in Images}
\label{sec:clustering}
For the input image sequence $\{I_i\}_{i=1}^{C}$, we identify characters and obtain visual character co-reference chains $\{\mathcal{V}_i\}_{i=1}^{K_v}$.
% \todo{We need to say that the we are now assuming that the bounding boxes are faces, not whole person as in the GS annotation! Also, how do we get the face bounding boxes in (1)?}

We experiment with two different approaches to obtaining face regions and features: (1) the pre-trained face detector MTCNN \cite{xiang2017joint} in conjunction with Inception Resnet \cite{szegedy2017inception} pretrained on VGGFace2 \cite{cao2018vggface2} for feature extraction, and (2) MTCCN for face detection combined with the CLIP pre-trained vision-language model \cite{radford2021learning} for feature extraction.

After obtaining the face features, we employ k-means \cite{lloyd1982least} on the face features to obtain the co-reference chains.
k-means requires us to specify $K$, the number of clusters, before we run the algorithm. We experiment with $2\leq k \leq 10$, as we assume a maximum of ten characters in a story and use the Calinski and Harabasz metric \cite{calinski1974dendrite} to evaluate the visual co-reference results, and select the $k$ with the highest score as the final cluster number $K_v$ for the image sequence $\{I_i\}_{i=1}^{C}$.
Note that some clustering algorithms (e.g., mean-shift) don't require a predefined number of cluster. But in our experiments k-means outperforms all other algorithms we tried (details in the supplementary material).

% \paragraph{CLIP for Face Similarity}
% \todo{This paragraph seems pretty obvious; maybe we can delete it to save space.}
% CLIP \cite{radford2021learning} is pretrained on a variety of image-text pairs to learn a shared semantic space for vision and language, enabling it to compute the similarity between images and text.
% However, CLIP can also be used for computing image similarities.
% Specifically, we feed the detected face regions into CLIP image encoder to obtain the face features for clustering.

\subsection{Character Grounding}
\label{alg:align}
We aim to align textual co-reference chains $\{\mathcal{T}_i\}_{i=1}^{K_t}$ and visual co-reference chains $\{\mathcal{V}_i\}_{i=1}^{K_v}$ at this stage.
We model character grounding as a bipartite graph matching problem.
Specifically, we propose to first generate a similarity matrix $A$ of shape $(K_t, K_v)$, where each element $A_{ij}$ represents the similarity of textual chain $\mathcal{T}_i$ and visual chain $\mathcal{V}_j$.
Then we apply the Kuhn–Munkres algorithm, also known as the Hungarian algorithm, to the similarity matrix and obtain the matching results.
We propose two methods for computing the similarity between $\mathcal{T}_i$ and $\mathcal{V}_j$, namely using distributional similarity and using pre-trained vision-language model.

\paragraph{Distributional Similarity}
Algorithm \ref{alg:cap} shows the details of the proposed distributional similarity-based alignment algorithm.
The assumption is that the textual mentions and visual appearances of the same character should have a similar distribution across the five sentences and images in a story.
For example, textual mentions in ${S_1}$ and ${S_2}$ and visual appearances in the corresponding images ${I_1}$ and ${I_2}$ tend to refer to the same character.

\begin{algorithm}[tb]
\caption{Distributional Similarity-based Alignment}
\label{alg:cap}
\textbf{Input}: Textual chains $\{\mathcal{T}_i\}_{i=1}^{K_t}$ and visual chains $\{\mathcal{V}_i\}_{i=1}^{K_v}$\\
\textbf{Output}: Alignment results $\mathcal{R}$.
\begin{algorithmic}[1] %[1] enables line numbers
\STATE Let $\mathcal{R} = [\,]$ and ${A} =$ zero matrix with shape $(K_t, K_v)$.
\FOR{$\mathcal{T}_i$ in $\{\mathcal{T}_i\}_{i=1}^{K_t}$}
\STATE $t_i \gets$ $C$-dim binary vector of the dist. of $\mathcal{T}_i$
    \FOR{$\mathcal{V}_j$ in $\{\mathcal{V}_j\}_{j=1}^{K_t}$}
    \STATE $v_j \gets$ $C$-dim binary vector of the dist. of $\mathcal{V}_j$
    \STATE ${A}_{ij} = \frac{t_i \cdot v_j}{|t_i|_1\cdot |v_j|_1}$ \COMMENT{$|\mathbf{x}|_1=\sum_{r=1}^{N}x_r$ }
    \ENDFOR
\ENDFOR
\STATE $\mathcal{R} =$ \textit{Kuhn–Munkres}\,$(A)$
\STATE \textbf{return} Alignment results $\mathcal{R}$.
\end{algorithmic}
\end{algorithm}

Mathematically, we use a $C$-dimensional binary vector ($t_i$ and $v_i$) to represent the distribution of a textual or visual character. More precisely, $t_i[k]=1$ when the $i$-th textual character is in the $k$-th sentence, and $v_i[k]=1$ when the $i$-th visual character appears in the $k$-th image.

\paragraph{CLIP-based Similarity}
As described in the previous section, we use CLIP to measure the similarity of image-text pairs.
CLIP is designed to handle one-to-one similarity.
We take the average of the similarity of each mention in the textual chain $\{\mathcal{T}_i\}_{i=1}^{K_t}$ and each visual appearance in the visual chain $\{\mathcal{V}_i\}_{i=1}^{K_v}$ as the similarity of the two chains:
\begin{equation}
    A_{ij} = \frac{1}{|\mathcal{T}_i|\cdot|\mathcal{V}_j|}\sum_{m=1}^{|\mathcal{T}_i|}\sum_{n=1}^{|\mathcal{V}_j|}\mathrm{CLIP}(\mathcal{T}_i^m, \mathcal{V}_j^n)
\end{equation}
% \todo{What is the intuition? This is just an average, right? We should say that.}
where $A_{ij}$ is the average similarity between $\{\mathcal{T}_i\}_{i=1}^{K_t}$ and $\{\mathcal{V}_j\}_{j=1}^{K_v}$, $\mathcal{T}_i^m$ denotes the $m$-th mention in the textual co-reference chain $\mathcal{T}_i$, $\mathcal{V}_j^n$ denotes the $n$-th visual appearance in $\mathcal{T}_i$, and $|\cdot|$ means the length of a co-reference chain.

\paragraph{Plural and Group Characters}
We first ground individual characters as described so far. Then we deal with plural and group characters.
For each plural/group textual co-reference chain, we compute its similarity with each visual co-reference chain using the same similarity method used for singular characters.
We then select the best match plural/group textual character for each visual character.
If the matching score is higher than a threshold, the visual character will be added to the plural/group co-reference chain.

After the above steps, we can obtain multi-modal co-reference chains $\{\mathcal{M}_k\}_{k=1}^{K_m}$, where $\mathcal{M}_k = \{\mathcal{T}_i, \mathcal{V}_j\}$ if the textual character $\mathcal{T}_i$ and the visual character $\mathcal{V}_j$ are aligned.
% \todo{Unclear, what do we mean by "best alignment characters"?}
Note that the remaining uni-modal characters will also be included in $\{\mathcal{M}_k\}_{k=1}^{K_m}$. Therefore, $K_m$ is equal to the maximum of $K_t$ and $K_v$.

\subsection{Importance Ranking}
\label{sec:ranking}
Intuitively, the more important a character is, the more crucial it is to the plot, and the more often it will be mentioned in the story.
In Experiments section we analyse the relationship between character frequency and importance and find that the two factors are well correlated.
We therefore use count-based importance ranking, which means the importance of each character depends on the length of its multi-modal co-reference chain.
\begin{equation}
    \textit{Importance}(\mathcal{M}_k) = |\mathcal{M}_k| = |\mathcal{T}_i| + |\mathcal{V}_j|
\end{equation}
%%% We have said this serveral times before.
%where $\mathcal{T}_i$ and $\mathcal{V}_j$ are the textual and visual coreference chains included in the multimodal chain $\mathcal{M}_k$, and and $|\cdot|$ means the length of a co-reference chain.
In addition, we're also interested in the importance ranking task in a single modality setting, i.e., when the input is the story text or the image sequence only.
For text-only input, the importance of each character is $|\mathcal{T}_i|$ and for image-only input, it is $|\mathcal{V}_i|$.

% \begin{figure}[h!]
%   \centering
%   \includegraphics[width=\textwidth]{imgs/t_baseline.jpg}
%   \caption{The illustration of the text baseline, where the words that refer to the same character (i.e., a co-reference chain) are labeled with the same color. The importance of each characters depends on the number of occurrences.}
%   \label{fig:t-baseline}
% \end{figure}

% Figure \ref{fig:t-baseline} shows the text baseline model, which uses co-reference resolution tool to obtain the mentions that refer to the same character, and the importance of each character depends on the number of occurrences.
% The performance of the text baseline depends on the accuracy of the co-reference resolution.
% Currently we're using the NeuralCoref.

% \subsection{Vision Baseline}
% \begin{figure}[h!]
%   \centering
%   \includegraphics[width=\textwidth]{imgs/v_baseline.jpg}
%   \caption{The illustration of the visual baseline. Bounding boxes containing the same character are of the same color. The importance of each character depends on the number of their appearance in the image sequence.}
%   \label{fig:v-baseline}
% \end{figure}
% Figure \ref{fig:v-baseline} illustrates the visual baseline.
% The model takes the image sequence as input, then uses 

% \subsection{Multimodal Baseline}
% \begin{figure}[h!]
%   \centering
%   \includegraphics[width=\textwidth]{imgs/m_baseline.jpg}
%   \caption{The multimodal baseline.}
% \end{figure}

%% file: 4_b3_experiments.tex
\section{Experiments}
\label{experiments}
% In this section we will evaluate our model for detecting and grounding important characters.
% For each subtask, we briefly introduce the experimental setting, explain the evaluation metrics, and analyse the results.
% We evaluate the proposed model on the VIST-Character dataset.

\subsection{Character Detection}
We use the spaCy PoS tagger (https://spacy.io/api/tagger) to obtain the nouns in text and then identify the characters using WordNet \cite{miller1995wordnet}.
For images, we obtain the faces using MTCNN \cite{zhang2016joint} and resize them to 160×160.
Table \ref{tab:detc_in_text_and_img} shows the results of character detection.

\begin{table}[t]
    \centering
    \begin{tabular}{lcc}
    \hline
    Modality      & Precision & Recall \\\hline
    Textual Story  & 74.4\% & 90.3\%  \\
    Image Sequence & 40.5\%  & 69.1\% \\\hline   
    \end{tabular}%
    \caption{Results for character detection in textual story and image sequence on VIST-Character.}
    \label{tab:detc_in_text_and_img}
\end{table}

\paragraph{Evaluation Metric}
We use precision and recall to evaluate textual and visual character detection.
For textual character detection, a predicted character phrase is considered as a correct character detection when the head word of the noun phrase is the same as the gold-standard one.
For visual character detection, a predicted face region is considered as a correct character detection when the bounding box of the face is entirely inside of an annotated body bounding box.

\paragraph{Analysis}
Overall, recall scores are higher than precision scores for both modalities.
The precision of image-based character detection is only 40.5\% and the gap with recall is around 28.6\%.
The reason is that there are a lot of background characters unrelated to the story in the images.
Comparatively, text is less noisy, and we can observe better character detection performance than in images.
However, we found that using WordNet and the hypernym method to filter character words can cause false positives.
For example, \textit{white} is usually used to describe color, but its primary sense in WordNet is a person name.
Also, some character words like \textit{great-grandmother} are not included in WordNet.

\subsection{Character Co-reference}
For textual stories, we experiment with two different approaches to obtain co-reference chains: (1) NeuralCoref \cite{clark2016improving} and (2) SpanBERT \cite{joshi2020spanbert}.
For image sequences, we use pre-trained MTCNN for face detection and compare two methods for face feature extraction: (1) Inception ResNet pre-trained on VGGFace2 and (2) the pre-trained CLIP model. The face instances are then clustered using k-means to obtain visual co-reference chains. Table \ref{tab:cluster_sim} presents the character co-reference results for both modalities.

\begin{table}[t]
\centering
\begin{tabular}{lcccc}
\hline
\multicolumn{1}{l}{\multirow{2}{*}{Model}} & \multicolumn{2}{c}{B-Cubed} & \multicolumn{2}{c}{Exact Match} \\
\multicolumn{1}{c}{} & Precision & Recall & Precision & Recall \\ \hline
NeuralCoref & \textbf{70.2\%} & 66.6\% & 29.2\% & 32.4\% \\ 
SpanBERT & 66.6\% & \textbf{70.0\%} & \textbf{30.0\%} &  \textbf{33.2}\%\\ \hline
ResNet & \textbf{57.0\%} & 60.5\% & 10.7\% & 17.5\% \\
CLIP & 51.8\% & \textbf{60.8\%} & \textbf{25.8\%} & \textbf{23.8\%} \\ \hline
\end{tabular}
\caption{Results for character co-reference. For the textual models, the input is the character mentions detected by PoS + WordNet. For the visual models, face instances are obtained by MTCNN and k-means is used for clustering.}
\label{tab:cluster_sim}
\end{table}

% \begin{table*}[htbp]
% \centering
% \begin{tabular}{llcccc}
% \hline
% \multicolumn{1}{l}{\multirow{2}{*}{Modality}} & \multicolumn{1}{l}{\multirow{2}{*}{Model}} & \multicolumn{2}{c}{Partial Match} & \multicolumn{2}{c}{Exact Match} \\
% \multicolumn{1}{l}{} & \multicolumn{1}{c}{} & Precision & Recall & Precision & Recall \\ \hline
% Textual Story & NeuralCoref & 50.5\% & 47.2\% & 35.4\% & 28.9\% \\ \hline
% \multirow{2}{*}{Image Sequence} & MTCNN + ResNet & 34.9\% & 47.3\% & 10.7\% & 17.5\% \\
%  & MTCNN + CLIP & 35.8\% & 51.3\% & 25.8\% & 23.8\% \\ \hline
% \end{tabular}
% \caption{Results for character clustering. For the visual models, k-means is used for clustering.}
% \label{tab:cluster}
% \end{table*}

\paragraph{Evaluation Metric}
Each character is represented by a co-reference chain.
We evaluate the co-reference chains from two perspectives.
B-Cubed recall and precision (following \citealt{cai-strube-2010-evaluation}) indicate the average percentage of the correctly detected mentions in a chain.
In contrast, exact match precision and recall require that the two chains are exactly the same.

\paragraph{Analysis}
% \textbf{visual}
% We can observe that MTCNN-based deep neural models perform much better than the traditional HOG-based face detection algorithms.
% However, this improvement doesn't have a significant impact on the clustering results.

% first partial vs. exact
In general, B-Cubed scores are higher than exact match, which means that co-referred mentions can be successfully detected, but it is difficult to identify the whole chain correctly. The reason is twofold. First, the fact that we are working on multi-sentence stories increases the difficulty of textual co-reference resolution.
Second, the image sequences in the VIST dataset usually contain different views of the same character, which makes it hard to form accurate clusters, see Figure~\ref{fig:arch} for examples.

% then text vs. image
Comparing modalities, we observe that co-reference resolution in text performs better than in images across the board.
We attribute this to the redundant characters detected in images, i.e., characters that are depicted but do not play a role in the story (e.g., people in the background).

For text-based models, SpanBERT outperforms NeuralCoref in terms of exact match, while for visual models, CLIP outperforms ResNet. In further experiments, we will use SpanBERT and CLIP, respectively.
% There are two reasons for this.
% First, Inception ResNet is pretrained on VGGFace2 that contains only 3.31 million labelled images, while CLIP is pretrained on 400 million images with captions.
% So CLIP can learn better image features and can generalise better.

\subsection{Character Grounding}
We evaluate distributional similarity-based and CLIP-based alignment applied on the output of the SpanBERT and MTCNN + CLIP pipelines.
In addition, we report the result of the alignment algorithm on gold standard input to obtain an estimate of its performance without accumulated error.
The results are shown in Table~\ref{tab:ali}.
% \usepackage{multirow}
% \begin{table}[htbp]
% \centering
% \resizebox{.45\textwidth}{!}{%
% \begin{tabular}{|c|c|c|c|}
% \hline
% Input                          & Model              & Recall & Precision \\ \hline
% \multirow{2}{*}{Gold-standard} & Distribution-based & 29.8   & 25.1      \\
%                               & CLIP-based         &        &           \\ \hline
% \multirow{2}{*}{Predictions}   & Distribution-based & 76.4   & 77.5      \\
%                               & CLIP-based         &        &           \\ \hline
% \end{tabular}
% }
% \caption{The performances of the distribution similarity-based alignment algorithm with and without the ground truth annotations as input.}
% \label{tab:ali}
% \end{table}
\begin{table}[t]
\centering
\resizebox{.45\textwidth}{!}{%
\begin{tabular}{llcc}
\hline
Input                          & Model              & Recall & Precision \\ \hline
\multirow{2}{*}{Predictions} & Distribution-based & \textbf{27.3\%}   & 27.5\%    \\
                               & CLIP-based         & 21.7\%    & \textbf{28.5\%}          \\ \hline
\multirow{2}{*}{Gold-standard}   & Distribution-based & \textbf{77.5\%}   & -      \\
                               & CLIP-based         &  58.5\%      &  -         \\ \hline
\end{tabular}
}
\caption{Performance of distribution similarity-based and CLIP-based alignment with and without the gold standard annotations as input. Best results are highlighted in bold.}
\label{tab:ali}
\end{table}

\paragraph{Evaluation Metric}
We use recall and precision to evaluate the character grounding task.
Note that recall is preferred to precision in this task, as visual character detection returns redundant characters (see above), which depresses the precision score.

\paragraph{Analysis}
% We can observe that the accuracy of the distribution-based method with gold standard input reaches 77.5\%.

% first gold-standard vs. predictions
We can observe that the performance is much better with gold-standard input than when detection-and-co-reference is used.
This indicates that the alignment algorithm works well, but is very sensitive to errors in the input. The errors of previous components accumulate and have a negative impact on the alignment performance.

% CLIP vs. distribution
Interestingly, CLIP does not outperform the distribution-based model.
The reason is that CLIP is pre-trained on image-caption pairs whose the caption is usually specific and detailed.
However, the VIST stories contain many generic words (e.g., \textit{he}, \textit{boy}) that appear frequently in the textual co-reference chains, making it difficult for CLIP to predict the alignment relation.
See the supplementary material for more details on the CLIP-based grounding results.

\subsection{Importance Ranking}

Intuitively, there is a positive relationship between the importance of a character and and its frequency: characters that are central to the plot are mentioned more often than side characters that don't contribute much to the story.

To verify the assumption that character importance is related to frequency, we compute the Pearson's correlation between the number of occurrences of a character in a story (i.e., the length of gold-standard co-reference chain), and the importance rating assigned to that character by the annotators.
Specifically, we investigate the correlation between a character's importance rating and (1) the number of occurrences in story text only, (2) the number of occurrences in image sequence only, and (3) the number of occurrences in both the image sequence and associated story text. Note that the correlation coefficient is uncomputable when there is only one character or all characters appear equally often in a story, so these instances were removed.
The result is shown in Table \ref{tab:corr} and confirms our hypothesis that character importance is correlated with character frequency.

\begin{table}[t]
\centering
\begin{tabular}{lc}
\hline \text {Story Type} &\text {Pearson's Correlation} \\
\hline
\text{Textual Stories} & 0.61  \\
\text{Visual Stories}  & 0.55  \\
\text{Multi-modal Stories} & 0.62  \\
\hline
\end{tabular}
\caption{Pearson's correlation between the number of occurrences and the importance ranking in different story modalities in the gold-standard VIST-Character dataset.}
\label{tab:corr}
\end{table}

We now turn to the most important character in a story, the protagonist. Being able to identify the protagonist is particularly important for story understanding or generation, as the plot revolves around this character. In our setting, the protagonist is expected to be the character with the highest frequency. We can therefore predict the protagonist as the character with the longest co-reference chain. We again compare text-only, image-only and multi-modal settings.
We use SpanBERT for the text-only setting and use MTCNN + CLIP for the last two. The results are given in Table~\ref{tab:ranking}.

\begin{table}[t]
\centering
\begin{tabular}{lccc}
\hline
Model               & P@1 & P@3 & P@5 \\ \hline
Text-only       & 53.6\%    & 79.2\%    & 87.4\%    \\ %\hline
Image-only     & 30.0\%    & 30.8\%    & 31.3\%    \\ %\hline
Multi-modal  & 29.4\%    & 34.7\%    & 34.3\%    \\ \hline
\end{tabular}
\caption{Performance for identifying the most important characters of a story. P@k means Precision@k. P@1 corresponds to identifying the protagonist. Multi-modal means that both the text and the images of a story are used.}
\label{tab:ranking}
\end{table}

\paragraph{Evaluation Metric}
We use the metric Precision@k to measure performance in predicting the most important characters. Precision@1 corresponds to predicting the protagonist correctly, while Precision@3 or Precision@5 indicate how many of the three or five most important characters have been identified correctly, compared to the human ratings of character importance in our VIST-Character dataset.
Note that for the multi-modal model, the precision score is calculated based on the match of multi-modal co-reference chains. Two multi-modal co-reference match if the head words (first textual mention) are the same.

\paragraph{Analysis}
The text-based model achieves the best performance among the three settings with Precision@5 reaching 87.4\%.
The reason is that the story text is treated as the evidence based on which annotators have rated character importance in our dataset. Moreover, the noise in the story text is less than in the image sequence, which makes the text-based importance ranking less error-prone than ranking based on images.

The results of the image-only model are significantly worse, with precision scores of only around 30\%. The reasons are two-fold. Firstly, the performance of clustering for co-reference resolution is not sufficiently high and thus adds a lot of noise to the downstream ranking task.
Secondly, the current evaluation method only considers the head image of the visual character co-reference chain.
Therefore, better co-reference resolution methods and more reasonable evaluation methods are needed to improve the image-only setting.

We can observe that the multi-modal setting performs slightly better than visual setting in terms of Precision@3 and Precision@5.
This means that the injection of textual evidence helps improve the performance of the image-only model somewhat.
From another perspective, the injection of visual evidence degrades the performance of the text-only model as the redundant characters in images bring a lot of noise to the length of the multi-modal co-reference chains if aligned with the wrong textual characters.
But we don't think the current method realizes the full potential of the multi-modal model given the large gap between the performance of text-only and multi-modal settings.

\subsection{Plural and Group Characters}
\begin{table}[]
\centering
\begin{tabular}{lcccc}
\hline
\multirow{2}{*}{Input} & \multicolumn{2}{c}{Grounding} & \multicolumn{2}{c}{Ranking} \\
                       & Precision       & Recall      & P@1     & P@5     \\ \hline
Singular          & 34.5\%          & 21.5\%      & 33.2\%       & 32.4\%       \\
Plural            & 20.9\%          & 47.5\%      & -            & -            \\ \hline
Both                   & 27.3\%          & 27.5\%      & 27.9\%       & 37.9\%       \\ \hline
\end{tabular}
\caption{The character grounding and importance ranking performance of multi-modal input with and without plural and group characters. P@k means Precision@k.}
\label{tab:rank_pl}
\end{table}

Table~\ref{tab:rank_pl} shows the character grounding and importance ranking performance for multi-modal input with and without plural and group characters.
For character grounding, the precision of plural and group characters reaches 47.5\% while the recall is only 20.9\%.
The reason is our grounding algorithm sets a high threshold (i.e., 0.6) for linking a visual character to a plural or group textual character.
This threshold prevents the model from overpredicting plural characters, given that these are relatively rare in the VIST stories.

For important ranking, we observe that Precision@1 drops by five points with plural and group characters, while Precision@5 increases by about five points.
The reason is that the protagonist is not a plural or group character in most cases.
Therefore, including plural and group nouns makes the most important character prediction more error prone.
However, the top five characters usually contain plural or group characters, thus Precision@5 will be higher when we consider plural and group characters.

%% file: 5_related.tex
\section{Related Work}

% The proposed task of character detection and grounding in visual stories contains several subtasks: co-reference resolution, character detection and clustering in images, character grounding, and character ranking.
% Next we review the related work in these research areas.
\paragraph{Visual Storytelling}
Visual storytelling was first proposed by \citet{huang2016visual} and has generated a lot of interest since, starting with early work \cite{gonzalez2018contextualize, kim2018glac} that employed a simple encoder-decoder structure, with a CNN to extract visual features and an RNN to generate text. More recent visual storytelling approaches \cite{xu2021imagine,chen2021commonsense,hsu2020knowledge,yang2019knowledgeable} introduce external knowledge to give models the necessary commonsense to reason. Sometimes, scene graphs are used to model the relations between objects \cite{lu2016visual,hong2020diverse,wang2020storytelling}.
However, none of the existing approaches explicitly consider character information -- characters are just treated like other objects. This results in a coherent sequence of events rather than a story with a clear plot driven by strong characters. Our work makes it possible to address this limitation by training character-centric models.

\paragraph{Detecting Important People in Images}
There is some prior work on detecting important people in single still images.
\citet{li2018personrank} contribute the multi-scene dataset and an NCAA Basketball dataset for important people detection in single images.
\citet{li2019learning} propose a deep importance relation network that learns the relations between the characters in an image to infer the most important person. Our work deals with image sequences, making important person detection more difficult is requires re-identification and co-reference resolution.

\paragraph{Character-grounded Video Description}
There is a long line of work which aims to link character mentions in movie or TV scripts with their visual tracks \cite{yu2020character,rohrbach2017generating,bojanowski2013finding,tapaswi2012knock,everingham2006hello,sivic2009you}. Important datasets for this task include ActivityNet Captions \cite{caba2015activitynet} augmented with bounding boxed and noun phrase grounding \cite{zhou2019grounded}.
Our plural and group character annotation is inspired by their annotation guidelines.
The M-VAD Names dataset \cite{pini2019m} extends the Montreal Video Annotation Dataset \cite{torabi2015using} with character names and corresponding face tracks, but does not include pronouns or plural nouns on the language side.
\citet{rohrbach2017generating} propose the MPII-MD Co-ref+Gender dataset, which contains the co-reference and grounding information for singular characters.
Compared with this work, our dataset includes plural/group character annotations and richer visual information in terms of the number of bounding boxes (4,989 vs. 2,649). Also, none of the existing datasets considers the importance of characters, making it difficult to distinguish protagonists from side characters. In modeling terms,
\citet{rohrbach2017generating} handle video description jointly with character grounding and co-reference and 
\citet{yu2020character} learn visual track embeddings and textualized character embedding to achieve character grounding and re-identification.
%However, these works ignore the plural and group characters due to the limitation of dataset.
%In another hand, it is difficult to identify or track the same character across discrete and discontinuous images.

\paragraph{Character-Centred Story Research}
This line of work aims to understand how characters drive the development of a story, resulting in models of character detection and persona learning
\cite{bamman2013learning,bamman2014bayesian,vala2015mr,flekova2015personality} and model of the interactions between characters \cite{iyyer2016feuding,srivastava2016inferring,chaturvedi2017unsupervised,kim2019frowning}.
\citet{surikuchi2019character} extract and analyse the character mentions from the VIST dataset.
Our approach to character mention detection is inspired by their work.
Other work has modeled the emotional trajectory of the protagonist \cite{brahman-chaturvedi-2020-modeling}. The character-centric storytelling model of
\citet{liu2020character} uses character-embeddings to guide the generation process. \citet{brahman-etal-2021-characters-tell} present a dataset of literary pieces paired with descriptions of the characters in them. 
In contrast to our work, all these approaches work on text only.

%% file: 6_conclusion.tex
\section{Conclusion}
We introduced the VIST-Character dataset, which extends the test set of VIST with co-reference chains for textual and visual character mentions and their importance rating.
Utilizing this dataset, we proposed the important character detection and grounding task, which requires character detection, co-reference, visual grounding and ranking.
We proposed two simple, unsupervised models for this task: one using distributional similarity and one based on the large pre-trained vision-language model CLIP.
In the future, we will exploit the proposed models to build a character-centred visual storytelling model, and extend it to longer and complicated narratives like comic books or movies.
We will release the dataset and codebase of the project, and hope this will benefit work in character-centric story understanding and generation.